\documentclass[10 pt, conference]{article}  

\usepackage{epsfig}
\usepackage{graphicx}
\usepackage{amsmath, amsfonts, amssymb, gensymb}
\usepackage{booktabs}
\usepackage{algorithm}
\usepackage{algorithmic}
\usepackage{makecell}
\usepackage{multirow}
\usepackage{flushend}
\usepackage[pagebackref=true,breaklinks=true,letterpaper=true,colorlinks,bookmarks=false]{hyperref}
\usepackage{SCITEPRESS}     

\newcommand{\fs}{\text{.} }
\newcommand{\com}{\text{,} }
\renewcommand{\L}{\mathcal{L}}

\newcommand{\etal}{\textit{et al. }}

\begin{document}
\title{\LARGE \bf
UAV-ReID: A Benchmark on Unmanned Aerial Vehicle Re-Identification in Video Imagery
}

\author{Anonymous Authors}

\author{Daniel Organisciak$^{1}$, Matthew Poyser$^{2}$, Aishah Alsehaim$^{2}$, Shanfeng Hu$^{1}$, Brian K. S. Isaac-Medina$^{2}$, Toby P. Breckon$^{2}$, Hubert P. H. Shum$^{2*}$

\affiliation{$^{1}$ Department of Computer and Information Sciences, Northumbria University, Newcastle upon Tyne, UK}
\affiliation{$^{2}$ Department of Computer Science, Durham University, Durham, UK}
\affiliation{$^{*}$ Corresponding author: hubert.shum@durham.ac.uk}}

\abstract{
As unmanned aerial vehicles (UAV) become more accessible with a growing range of applications, the risk of UAV disruption increases. Recent development in deep learning allows vision-based counter-UAV systems to detect and track UAVs with a single camera. However, the limited field of view of a single camera necessitates multi-camera configurations to match UAVs across viewpoints -- a problem known as re-identification (Re-ID). While there has been extensive research on person and vehicle Re-ID to match objects across time and viewpoints, to the best of our knowledge, UAV Re-ID remains unresearched but challenging due to great differences in scale and pose. We propose the first UAV re-identification data set, \emph{UAV-reID}, to facilitate the development of machine learning solutions in multi-camera environments. UAV-reID has two sub-challenges: \emph{Temporally-Near} and \emph{Big-to-Small} to evaluate Re-ID performance across viewpoints and scale respectively. We conduct a benchmark study by extensively evaluating different Re-ID deep learning based approaches and their variants, spanning both convolutional and transformer architectures. Under the optimal configuration, such approaches are sufficiently powerful to learn a well-performing representation for UAV (81.9\% mAP for Temporally-Near, 46.5\% for the more difficult Big-to-Small challenge), while vision transformers are the most robust to extreme variance of scale.
} 

\onecolumn \maketitle \normalsize \setcounter{footnote}{0} \vfill
\section{\uppercase{Introduction}}\label{sec:intro}

Unmanned aerial vehicles (UAV) are becoming more accessible and more powerful through technological advancement. Their small size and manoeuvrability allows for a wealth of applications, such as film-making, search and rescue, infrastructure inspection, and landscape surveying. However, 
the malicious or accidental use of UAVs could pose a risk to aviation safety systems or privacy. This necessitates the development of counter-UAV systems. Due to the recent development of computer vision and deep learning, vision-based UAV detection and tracking systems have become more robust and reliable \cite{issacmedina21unmanned, jiang2021antiuav}. 

There are two major issues with existing vision-based counter-UAV systems: firstly, many systems are only built for a single camera -- once a UAV leaves the range of capture, the captured information can no longer be re-used; secondly, to help prevent ID-switching and handle occlusion, many tracking frameworks rely on a generic re-identification (Re-ID) module, which cannot comprehensively handle the complex challenges that come with re-identifying UAVs \cite{issacmedina21unmanned}. 

Of these, DeepSORT \cite{deepsort} and Tracktor \cite{tracktor} are perhaps the two most prominent frameworks within the tracking domain. Tracktor requires the network to associate new and previously disassociated tracks. DeepSORT on the other hand, employs its Re-ID module at each time step within the Hungarian Algorithm \cite{hungarian} to associate new and old detections. Indeed, in the original and many subsequent works, the association metric is heavily weighted towards the output of the Re-ID network, especially when camera motion is particularly prevalent. The reliance upon robust reidentification networks by both single and multi-view tracking frameworks is evident and thus dedicated study to effectively re-identify UAVs is essential to solve both problems. To enable a cross-camera UAV system, effective Re-ID is needed to match observed UAVs from one camera to another from different angles, poses, and scales. Generic Re-ID mechanisms within off-the-shelf tracking frameworks can be improved by designing a bespoke UAV Re-ID system to handle these extreme changes.

\begin{figure}[t]
\centering
{\includegraphics[width=0.8\linewidth]{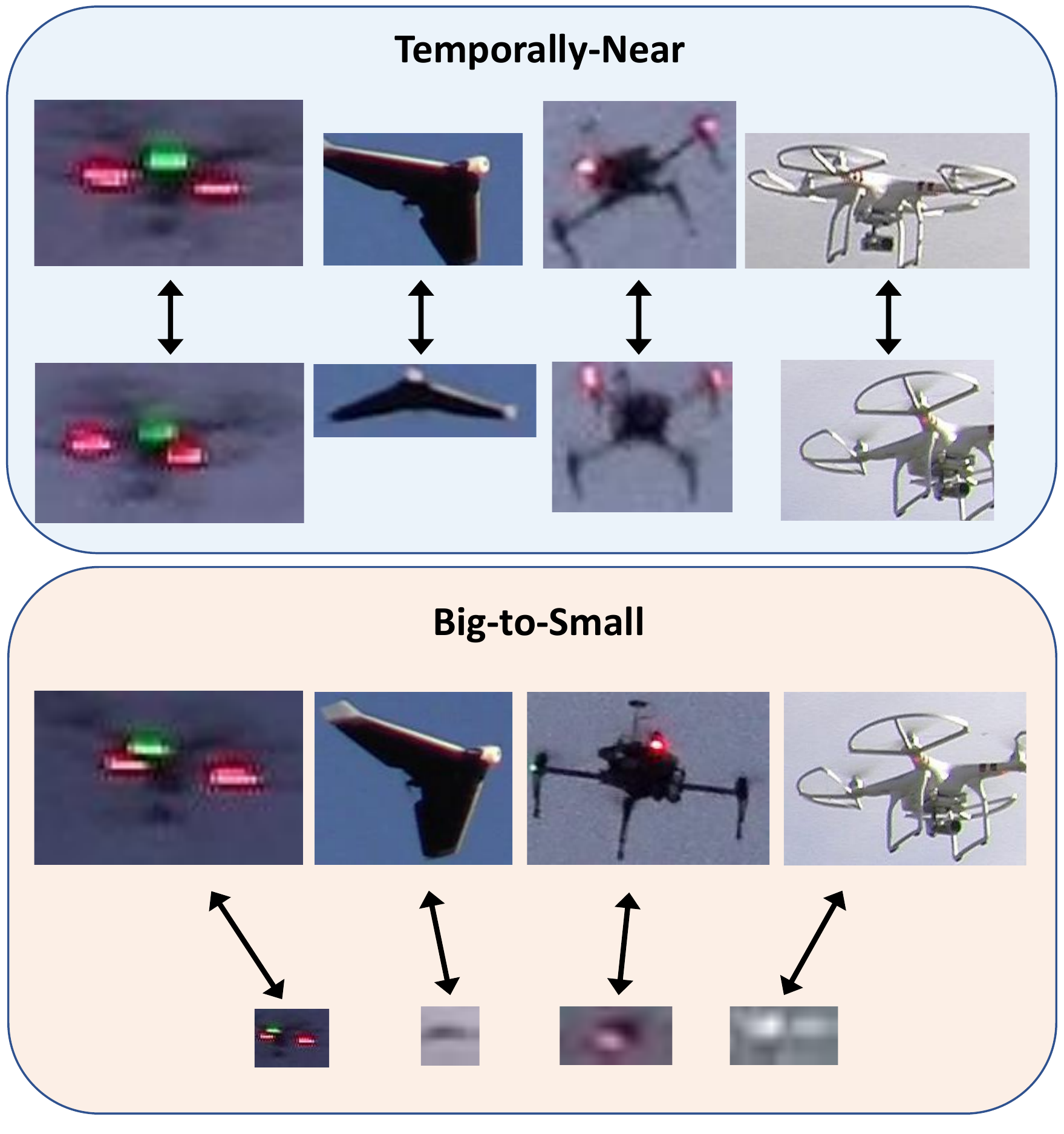}}
\caption{The two Re-ID sub-challenges we explore. Temporally-Near models the difficulties of tracking UAVs, whereas Big-to-Small simulates cross-camera or temporally distant challenges of matching UAVs.} \label{fig:Dataset}
\end{figure}

There has been a large body of research in Re-ID for pedestrians \cite{ye2021deep} and vehicles \cite{deng2021trends}. Most state-of-the-art person Re-ID research typically employ engineering solutions to improve performance, such as a `bag of tricks' \cite{luo2019bag}, which identifies several key Re-ID principles to adhere to. Indeed, such methods have been illustrated to introduce sufficient robustness such that state of the art results for person Re-ID can be achieved, even by shallow networks \cite{aishahnot3d}. Other works exploit the relatively static colour profile of pedestrians across views with part-based systems \cite{sun2018beyond, fu2019horizontal}. In contrast, vehicles have drastically different appearances across views, so this information must be incorporated into the model \cite{zhou2018aware}. For UAV Re-ID, even more consideration is required due to the increased potential for the change in viewing angle of the UAV target from any given camera position owing to their unconstrained motion in 3D space.

As a result of their unconstrained aerial motion UAV may undergo considerably greater changes in scale relative to the camera than comparable pedestrian or vehicle targets. Furthermore, they can appear from any angle on the sphere, compared to pedestrians and vehicles, that are typically captured from a 0-30\degree elevation. As a result of these extended inter-view object tracking challenges, a study is required to evaluate the performance of existing Re-ID systems on these challenges that UAVs provide.

However, to the best of our knowledge, there has been no research on UAV Re-ID. In the absence of a true multi-view UAV data set, we propose the \emph{UAV-reID} dataset, as a new and challenging benchmark for UAV Re-ID. To simulate Re-ID challenges, UAV-reID has two sub-challenge dataset splits: \emph{Temporally-Near} 
 aims to evaluate the performance across a short time distance, as Re-ID modules within tracking frameworks must successfully identify the same UAV in subsequent frames within videos; \emph{Big-to-Small} evaluates Re-ID performance across large scale differences. The results inform Re-ID performance of matching UAVs across two cameras, or across a large timescale within the same camera. Figure \ref{fig:Dataset} visualises these sub-challenges.

We conduct a benchmark study of state-of-the-art deep neural networks and frameworks designed for Re-ID, including ResNet \cite{he2016deep}, SE-ResNet \cite{hu2018squeeze}, SE-ResNeXt \cite{xie2017aggregated}, Vision Transformers (ViT) \cite{dosovitskiy2021image}, ResNetMid \cite{yu2017devil}, Omni-scale Network (OSNet) \cite{zhou2019omni}, Multi-level Factorisation Network (MLFN) \cite{chang2018multi}, Parts-based Convolutional Baseline (PCB) \cite{sun2018beyond}, Harmonious Attention Network (HACNN) \cite{li2018harmonious}, and Not 3D Re-ID (N3D-ReID) \cite{aishahnot3d}. We test all baselines with a cross-entropy loss, a triplet loss, a combined loss and a multi-loss.

Experimental results show that existing Re-ID networks cannot transfer seamlessly to UAV Re-ID, with the best setup achieving 81.9\% mAP under Temporally-Near and 46.5\% under Big-to-Small. ViT is the most robust to extreme scale variance. This compares to 84.61\% \cite{aishahnot3d} performance when evaluated on typical pedestrian or vehicle targets (e.g. MARS dataset \cite{zheng2016mars}) as are commonplace in existing Re-ID evaluation benchmarks.

The contributions of this paper are summarised as follows:
\begin{itemize}
    \item proposal of the novel task of UAV Re-ID to match UAVs across cameras and time frames, to improve visual security solutions on UAVs
    \item Construction of the first 
    UAV Re-ID data set 
    \emph{UAV-reID}, to facilitate Re-ID system development and benchmarking. This is formulated by two sub-challenge dataset splits, \emph{Temporally-Near} and \emph{Big-to-Small}, to evaluate performance under conditions where Re-ID is used in a practical environment, and remain applicable even when dataset availability is constrained.
    \item creation of the first extensive benchmark over a variety of state-of-the-art Re-ID architectures within the UAV domain: ResNet, SE-ResNet, SE-ResNeXt, ViT, ResNetMid, OSNet, MLFN, PCB, HACNN, N3D-ReID; with critical evaluation of their strengths and weaknesses, obtaining 81.9\% mAP on Temporally-Near and 46.5\% mAP on Big-to-Small. 
\end{itemize}

\begin{figure*}[t]
\centering
{\includegraphics[width=0.9\linewidth]{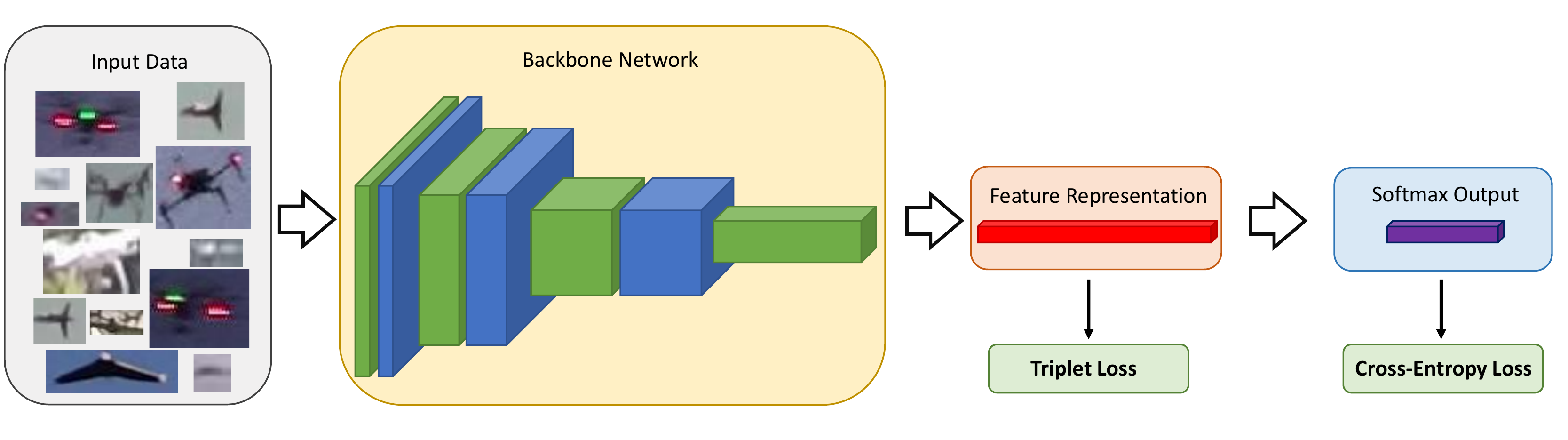}}
\caption{An overview of the pipeline for all of our experiments. Input data from the proposed UAV-ReID data set is processed by the given backbone network to obtain a feature representation. This feature representation is used in the triplet loss, and also goes through a softmax classification layer to be used in the cross-entropy loss. The backbone networks we evaluate are presented in Section \ref{ssec:backbones}} \label{fig:Pipeline}
\end{figure*}

\section{\uppercase{Related Work}}\label{sec:rw}
Here we detail existing literature with respect to evolution of Re-ID methodology, and its application within the UAV domain.
\subsection{Re-identification}
Before large-scale Re-ID data sets were proposed, traditional machine learning works focused on designing hand-crafted features and learning distance metrics \cite{KaranamSystematic}. 
Even though UAV-reID is a small data set, UAVs can appear at many different sizes and it is difficult to hand-craft features that are robust to this extreme scale transformation. 
For this reason, we conduct this study on deep learning methods which are capable of computing robust features \cite{he2016deep, hu2018squeeze} and demonstrate supreme performance on other Re-ID tasks \cite{sun2018beyond, hermans2017defense, li2018harmonious}.

Re-ID with deep learning became popular after the release of ResNet \cite{he2016deep} with many works taking advantage of the complex information that very deep features could encode. 
More recently, extensions such as SE-ResNet \cite{hu2018squeeze} and SE-ResNeXt \cite{xie2017aggregated} have seen more use as a generic backbone architecture for Re-ID frameworks.
These frameworks commonly consist of engineering solutions \cite{luo2019bag} for easier representation matching. Person Re-ID \cite{ye2021deep} frameworks typically take advantage of the similar colour profile of pedestrians across views, often by splitting the image into parts \cite{sun2018beyond, fu2019horizontal} to separately encode information of the head, clothes, and shoes.
Conversely, vehicle Re-ID \cite{deng2021trends} has to contend with shape information that undergoes significant deformation across viewpoints, which may require encoding viewpoint information within the model \cite{zhou2018aware, meng2020parsing}.

Compared to most classification problems, Re-ID often contains many classes (individuals, vehicles, UAVs) and few samples per class. This makes learning class-specific features difficult. To handle this problem, it is often beneficial to consider metric learning, usually in the form of the triplet loss \cite{hoffer2015deep} or centre loss \cite{wen2016discriminative}. The triplet loss in particular has seen extensive use for person  \cite{hermans2017defense, cheng2016person} and vehicle \cite{kuma2019vehicle} Re-ID, and can even handle both tasks simultaneously \cite{organisciak2020unifying}. Within this study it is therefore natural to consider the triplet loss for UAV Re-ID.

\subsection{Computer Vision on UAV}
A large body of research applying computer vision to imagery captured by UAVs has been developed, including object detection \cite{gaszczak11uavpeople}, visual saliency detection \cite{sokalski10uavsalient, gokstorp21saliency}, visual segmentation \cite{seg1}, target tracking \cite{target_tracking_1} and aerial Re-ID \cite{aerial_reid_1, aerial_reid_2, aerial_reid_3}. However, the study of such tasks where UAV are the main object of interest has not been extensively investigated. Most UAV-related computer vision research is focused on deep learning approaches for UAV detection and tracking \cite{issacmedina21unmanned, uav_yolo, craye2019spatio, airbone_visual_detection}. In this context, some data sets have been created to investigate novel visual-based counter-UAV systems. The Drone-vs-Bird Challenge data set \cite{coluccia2019drone} collects a series of videos where UAV usually appear small and can be easily confused with other objects, such as birds. Recently, the Anti-UAV data set \cite{jiang2021antiuav} has been proposed to evaluate several tracking algorithms in both optical and infrared modalities. Despite the advances in the counter-UAV domain and the available data sets, this study represents the first time UAV Re-ID has been investigated. We believe this is a crucial task for future vision-based counter-UAV systems, which are both passive in nature and, of course, afford visual confirmation of acquired UAV targets.

\section{\uppercase{Deep Neural Network Architectures}}\label{sec:method}
We present an overview of the deep learning architectures considered within this work in terms of both their underlying convolutional neural network backbone and the loss function that they employ for weight optimisation.

\subsection{Network Backbones} \label{ssec:backbones}
Deep neural networks (DNN) are machine learning systems that use multiple layers of non-linear computation to model the complicated relationship between the input and output of a problem. Convolutional neural networks (CNN) are particularly suited for image-based object identification and tracking in computer vision applications. Firstly, CNNs can capture object features irrespective of their spatial locations on an image, due to the shift-invariance of convolution kernels. Secondly, modern CNNs can detect objects of complex shapes, sizes, and appearance by stacking multiple convolution kernels to learn powerful feature representations. We describe a selection of state-of-the-art CNNs and generic Re-ID frameworks that we evaluate for UAV Re-ID. Our overall framework is shown in Figure \ref{fig:Pipeline}.

\textbf{ResNet:}
Residual neural networks \cite{he2016deep} are
a popular variant of CNNs that connect adjacent layers of a network (residuals) with an identity mapping. Learning residuals enables training significantly deeper architectures to obtain more powerful features. In our experiments, we use the 18-layer, 34-layer, and 50-layer configurations.

\textbf{SE-ResNet:} ResNets are powerful but can still be improved by learning and re-weighting the hidden convolutional feature maps using attention. The popular Squeeze-Excitation (SE) network \cite{hu2018squeeze} introduces a channel attention mechanism to identify and appropriately weight important feature maps.


\textbf{SE-ResNeXt:} Another line of improvement for ResNet is ResNeXt \cite{xie2017aggregated}, which maintains the identity skip connection while splitting the feature mapping of each layer into multiple branches. This increased dimension of network representation power has shown to be more effective for image recognition and object detection. 

\textbf{ViT:} Transformers have recently become ubiquitous in natural language processing. Motivated by this, Dosovitskiy \etal \cite{dosovitskiy2021image} migrated transformers into computer vision to propose \emph{Vision Transformers} (ViT). This architecture learns the relationship among all image patches for downstream tasks. We evaluate ViT with image patches of size $16 \times 16$ with the `small' (8-layer) and `base' (12-layer) configurations.


\textbf{ResNet50-mid:} A common practice of image representation learning in computer vision is to take hidden features from the penultimate CNN layer as image embeddings. Yu \etal \cite{yu2017devil} explore fusing embeddings from earlier layers to improve the performance of cross-domain image matching. Fusing representations from different layers has proven successful for other computer vision tasks on small objects \cite{liu2020deep}, highlighting its potential within UAV Re-ID systems.

\textbf{OSNet:} There have also been CNN architectures specifically designed for object Re-ID. Zhou \etal \cite{zhou2019omni} propose an omni-scale network, which improves Re-ID performance by learning to fuse features of multiple scales within a residual convolutional block. Each stream in the block corresponds to one scale to learn and the outputs of all streams are dynamically combined to create omni-scale features. Considering the expansive array of scales at which UAV can appear, OSNet is well-suited to the UAV Re-ID challenge.

\textbf{MLFN:} Multi-level Factorisation Network \cite{chang2018multi} is similar to OSNet in that it tries to capture discriminative and view-invariant features at multiple semantic levels. Unlike OSNet however, it composes multiple computational blocks, each containing multiple factor modules and a selection gate to dynamically choose the best module to represent the input. 

\textbf{PCB:} Different from holistic feature learning, Sun \etal \cite{sun2018beyond} propose a \emph{parts-based convolutional baseline} (PCB), which uniformly splits each input image into multiple parts. As the appearance consistency within each part is usually stronger than between parts, it proves easier to learn more robust and discriminative features for person Re-ID. A part pooling module is added to deal with outliers.

\textbf{HACNN:} Li \etal \cite{li2018harmonious} propose a \emph{harmonious attention network}, which tackles the challenge of matching persons across unconstrained images that are potentially not aligned. HACNN uses layers that incorporate hard attention, spatial attention and channel attention to improve person Re-ID performance on unconstrained images. We reformulate this system towards re-identification of UAV objects to thus enable evaluation of its performance within the counter-UAV domain

\textbf{The N3D-ReID Framework:}
The use of Re-ID best practices \cite{luo2019bag} alongside simple networks have been demonstrated to be a suitable replacement for more complex Re-ID networks, as identified by the Not 3D Re-ID Framework \cite{aishahnot3d} (N3D-ReID). By introducing a Batch Normalisation Neck between the deep backbone network and a multi-loss function explained in Section \ref{sssec:multiloss}, the authors were able to achieve state of the art results within the person Re-ID domain. Moreover, they utilize an additional backbone architecture denoted ResNet50-IBN-a \cite{ibnnet}, which introduces both batch normalisation \cite{batchnorm} and instance normalisation \cite{instancenorm} into the backbone architecture itself. As such, we further evaluate the performance of ResNet50-IBN-a and the backbone architectures outlined in Section \ref{ssec:backbones} within this separate re-identification framework in addition to that illustrated in Figure \ref{fig:Pipeline}.  All implementation details remain unchanged from the original paper \cite{aishahnot3d}. 

\subsection{Loss Functions}
In order to perform learning via weight optimisation across the specified deep neural network architecture, a loss function denoting relative network weight performance on the specified task is minimised via computational optimisation with corresponding weight updates via backpropagation. We detail a number of such loss functions which are considered within this study for the application of UAV Re-ID.

\textbf{Cross-Entropy Loss:}
The cross-entropy (CE) loss function is the standard loss that is used in most machine learning classification tasks. The negative log-likelihood between the true class labels and predicted class labels is minimised:
\begin{equation}
    \L_{\mathrm{CE}} = -\sum_{x \in \mathcal{X}} y_x \log f(x; \theta)\com
\end{equation}
where a network $f$ with parameters $\theta$ predicts the class of an input $x$ with a true class index $y_x$.

\textbf{Triplet Loss:} The triplet loss is a metric learning technique that decreases the distance between positive pairs of images and increases the distance of negative pairs. Metric learning is commonly used in applications such as verification and Re-ID, where there are many classes and few instances per classes. Because of the lack of class-specific data, the network cannot reliably learn class-specific information. The network instead learns to place images onto a manifold with similar images placed close to one another.

We denote a triplet, $t=(x, x^+, x^-)$, where $x$ is the query image, $x^+$ is an image of the same object, and $x^-$ is an image of a different object. The triplet loss function is formulated as follows:
\begin{equation}
\begin{split} \label{eq:triplet}
\mathcal{L}_{\mathrm{triplet}} = &\sum_{t\in \mathcal{T}} \max ( (||f^*(x; \theta)-f^*(x^+; \theta)||^2 \\ &- \, ||f^*(x; \theta)-f^*(x^-; \theta)||^2 + \alpha), 0 ) \mathrm{,}
\end{split}\end{equation}
where  $\mathcal{T}$ is the set of mined triplets, $||\cdot||^2$ is the Euclidean distance, and the feature representation $f^*(x; \theta)$ is obtained by passing input $x$ through network $f$ with parameters $\theta$, and taking the representation before the softmax classification layer. Negative images are pushed away from positive images by a margin of $\alpha$.

Triplets need to be sufficiently difficult in order to improve the performance of the model \cite{hermans2017defense}. We employ \emph{hard negative mining} to each query image in the batch. This means that within each iteration, the most difficult negative samples are considered and processed by the loss function. In turn, these samples maximise how much is learnt during backpropagation. Given a query image $q$, the hardest negative image in the gallery is found via $
    \min ||f^*(q) - f^*(g_i)||^2 \com $
where $g_i, i \in \{1, \ldots, B\}$ are the gallery images, $B$ is the batch size, and $||\cdot||^2$ is the Euclidean distance.

\textbf{Combined Loss:} In many Re-ID works, combining the two losses can lead to performance gains \cite{luo2019bag}. We test this setting for UAVs where both losses receive equal weight:
\begin{equation}
\L = \L_{\mathrm{CE}} + \L_{\mathrm{triplet}}\fs
\end{equation}

\textbf{Multi-Loss:} \label{sssec:multiloss}
Following the success of N3D-ReID \cite{aishahnot3d}, we further evaluate the performance of a multi-loss function that has demonstrated superior performance to more well-established loss functions within the person Re-ID domain. This loss is formulated as a weighted sum across cross-entropy loss, $\L_{ID}$, ranked list loss, $\L_{RLL}$, centre loss, $\L_\mathrm{centre}$, and erasing-attention loss, $\L_{E\_att}$, as follows:
\begin{equation}
\L = \L_{ID} + \L_{RLL} + \beta\cdot \L_\mathrm{centre} + \L_{E\_att}\fs
\end{equation}

As such, all losses receive equal weighting other than centre loss which serves to support $\L_{RLL}$, and thus receive weight $\beta$. We define $\L_{ID}$ as cross-entropy loss with additional Label Smoothing \cite{labelsmoothing}. $\L_{RLL}$ can be considered a direct alternative to triplet loss, and learns a hypersphere for each class additionally to triplet loss behaviour. Learning the hypersphere helps avoid intra-class data distribution that might be apparent within triplet loss, and particularly impactful when training with limited data. Finally, $\L_{E\_att}$ introduces additional attention to image samples that receive erasing under random erasing augmentation \cite{zhong2020random} such that its impact is increased, as implemented in \cite{aishahnot3d, reaimplementation}. This is particularly important when data availability is constrained so the effects of over-fitting are minimised during training; learning will be maximised from features extracted from erasing-augmented images that are less likely to contribute to UAV regions.

\begin{figure}[t]
{\includegraphics[width=\linewidth]{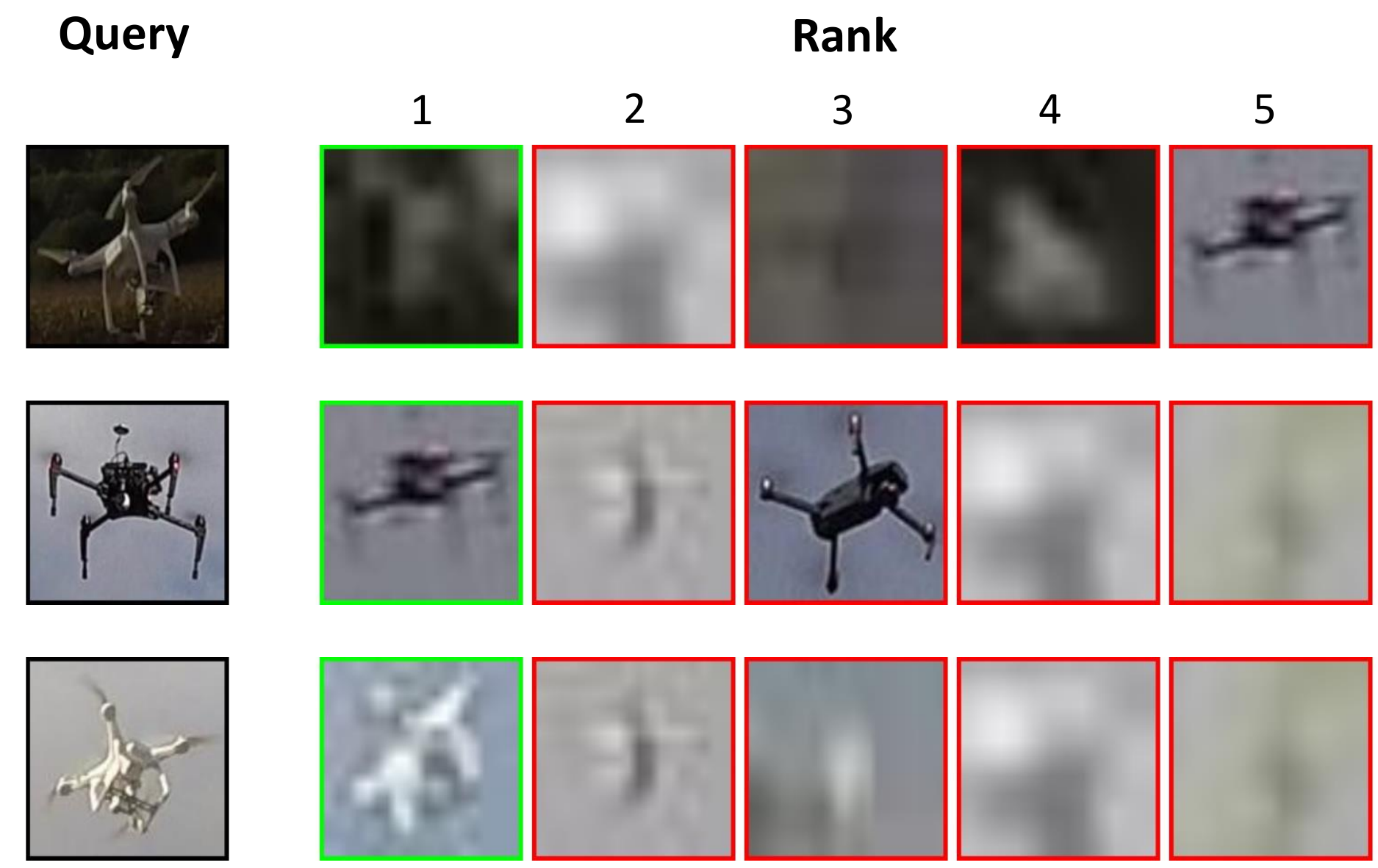}}
\caption{Examples from ViT with a combined loss on Big-to-Small. A green box indicates a correct Re-ID. ViT can extract salient features from very low-resolution images to  match UAVs across scale. } \label{fig:Ranks}
\end{figure}

\section{\uppercase{UAV Re-ID Dataset}}\label{sec:exp}
We present our dataset for the UAV Re-ID task and corresponding experimental setup.
\subsection{Data}
UAV-reID is designed to evaluate two practical applications of Re-ID. All data set instances are constructed via sampling from 
61 UAV videos. UAVs are cropped from single frames of these videos depending on the specific challenge. UAV images are then resized to size $224 \times 224$ Images are augmented via random flipping, random cropping, and random erasing \cite{zhong2020random}. Similar to early person Re-ID data sets, we include two images per identity for each setting. Across both challenges, our dataset contains 61 UAV identities and 244 UAV images.

We use 30 identities for training and the remaining 31 identities for testing. Our code can be found at \url{https://github.com/danielorganisciak/UAVReID}.


\subsection{Challenges}
\textbf{Temporally-Near:} \label{sssec:temporally_near}
Given a UAV video with $t$ frames, we consider UAVs in frames $\frac{t}{5}$ and $\frac{2t}{5}$. This temporal distance is close enough that UAVs remain at a similar size in most cases, but far enough for UAVs to appear from a different viewpoint. This simulates the task that a Re-ID module embedded within a tracking framework must perform, whereby UAVs undergo a limited transformation.

\textbf{Big-to-Small:} \label{sssec:big_to_small}
We obtain the largest and smallest UAV detections across the whole video. This simulates the task of matching known UAVs (for which we have rich visual information) with UAVs detected from a long distance. As such, we can identify the far-off UAV, and whether it poses a potential threat.

\renewcommand\arraystretch{1.2}
\begin{table*}[t]
\small
\centering
\caption{Methods Tested on the `Temporally-Near' sub-challenge.} \label{tab:TempNear}
\begin{tabular}{l|c|ccc|ccc|ccc}
\hline\hline
Backbone & ReID-Specific & \multicolumn{3}{c|}{CE} & \multicolumn{3}{c|}{Triplet} & \multicolumn{3}{c}{CE + Triplet} \\ \hline
 & & mAP & rank-1 & rank-5 & mAP & rank-1 & rank-5 & mAP & rank-1 & rank-5\\\hline
ResNet-18 & $\times$ & {\color{red} \bf 81.9} & {\color{red}\bf 77.4} & 77.4 & 72.7 & 61.3 & {\textit{\color{blue}74.2}} & 71.7 & 58.0 & 77.4 \\
ResNet-34 & $\times$  & 77.1 & 70.1 & 74.2 & 74.6 & {\color{red}71.0} & 71.0 & 74.4 & 61.3 & {\color{red}83.9} \\
ResNet-50 & $\times$  & 75.9 & 71.0 & 71.0 &{\textit{\color{blue} 75.5}} & {\color{red}71.0} & 71.0 & 76.7 & 67.7 & 77.4 \\
SE-ResNet-50 & $\times$  & 77.1 & 71.0 & 80.6 & 74.1 & 67.7 & {\textit{\color{blue}74.2}} & 79.4 & 74.2 & {\textit{\color{blue}80.6}} \\
SE-ResNeXt-50 & $\times$  & 75.8 & 71.0 & 77.4 & 66.8 & 61.3 & 64.5 & 76.2 & 74.2 & 74.2 \\
ViT Small & $\times$  & 75.6 & 67.7 & 74.2 & 74.1 & 64.5 & {\textit{\color{blue}74.2}} & 75.6 & 64.5 & 74.2 \\
ViT Base & $\times$  & 79.2 & {\textit{\color{blue}74.2}} & 77.4 & 73.2 & 67.7 & {\textit{\color{blue}74.2}} & {\textit{\color{blue} 81.3}} & {\color{red}\bf 77.4} & {\textit{\color{blue}80.6}}\\\hline
ResNet50mid &\checkmark  & 78.0 & 71.0 & {\color{red}\bf 87.1} & 74.0 & 67.7 & {\textit{\color{blue}74.2}} & 76.1 & 67.7 & 77.4 \\
OSNet &\checkmark  & 71.0 & 61.3 & 70.1 & 73.8 & 67.7 & 71.0 & 75.7 & 71.0 & 71.0\\
MLFN &\checkmark  & 69.9 & 61.3 & 71.0 & 73.4 & 67.7 & 67.7 & 65.7 & 58.1 & 61.3\\
PCB &\checkmark  & {\textit{\color{blue} 80.8}} & {\textit{\color{blue}74.2}} & {\color{red}\bf 87.1} & 73.2 & 67.7 & 67.7 & {\color{red}81.4} & {\color{red}\bf 77.4} & {\textit{\color{blue}80.6}}\\
HACNN &\checkmark  & 72.1 & 64.5 & 71.0 & {\color{red}77.7} & {\color{red}71.0} & {\color{red}77.4} & 74.5 & 64.5 & 77.4\\
\hline\hline                      
\end{tabular}
\\\medskip
Bold denotes the highest values in the table, red denotes the highest in each column, blue denotes the second highest in each column 
\end{table*}
\renewcommand\arraystretch{1}

\renewcommand\arraystretch{1.2}
\begin{table*}[h]
\small
\centering
\caption{Methods Tested on the `Big-to-Small' sub-challenge.} \label{tab:BigtoSmall}
\begin{tabular}{l|c|ccc|ccc|ccc}
\hline\hline
Backbone & ReID-Specific & \multicolumn{3}{c|}{CE} & \multicolumn{3}{c|}{Triplet} & \multicolumn{3}{c}{CE + Triplet} \\ \hline
 & & mAP & rank-1 & rank-5 & mAP & rank-1 & rank-5 & mAP & rank-1 & rank-5\\\hline
ResNet-18 & $\times$ & 40.3 & {\textit{\color{blue}32.3}} & 41.9 & 36.9 & 25.8 & 32.3 & 37.5 & 25.8 & 32.3 \\
ResNet-34 & $\times$  & 33.7 & 22.6 & 29.0 & 37.9 & 29.0 & 35.5 & 38.8 & 25.8 & 35.5 \\
ResNet-50 & $\times$  & 37.8 & 22.6 & {\textit{\color{blue}51.6}} & 39.0 & 29.0 & 35.5 & 42.9 & 29.0 & 35.5 \\
SE-ResNet-50 & $\times$  & 38.0 & 25.8 & {\textit{\color{blue}51.6}} & {\textit{\color{blue}42.5}} & 29.0 & 45.0 & 41.4 & 29.0 & 38.7 \\
SE-ResNeXt-50 & $\times$  & 40.0 & 29.0 & 35.5 & 31.9 & 16.1 & 29.0 & 38.8 & 29.0 & 32.3 \\
ViT Small & $\times$  & {\color{red}43.1} & {\color{red}35.5} & 35.5 & 39.0 & 22.6 & {\textit{\color{blue}41.9}} & 40.9 & 29.0 & 38.7 \\
ViT Base & $\times$  & 40.5 & 29.0 & {\color{red}\bf 54.8} & 36.9 & 22.6 & 32.3 & {\color{red}\bf 46.5} & {\color{red}\bf 35.5} & {\color{red}45.2}\\\hline
ResNet50mid &\checkmark  & 38.4 & 25.8 & {\textit{\color{blue}51.6}} & 42.3 & {\color{red}32.3} & 32.3 & {\textit{\color{blue}43.2}} & {\textit{\color{blue}32.3}} & 38.7 \\
OSNet &\checkmark  & 38.0 & 25.8 & 35.5 & 34.5 & 19.4 & 35.5 & 33.2 & 19.4 & 32.3\\
MLFN &\checkmark  & 38.1 & 22.5 & 38.7 & 36.8 & 25.8 & 32.3 & 33.9 & 22.6 & 25.8\\
PCB &\checkmark  & {\textit{\color{blue}41.3}} & {\textit{\color{blue}32.3}} & 35.5 & {\color{red}43.7} & {\color{red}32.3} & {\textit{\color{blue}41.9}} & 38.2 & 25.8 & 32.3\\
HACNN &\checkmark  & 36.0 & 19.4 & 45.2 & 39.4 & 25.8 & 32.3 & 41.2 & 25.8 & {\textit{\color{blue}41.9}}\\
\hline\hline                      
\end{tabular}
\\\medskip
Bold denotes the highest values in the table, red denotes the highest in each column, blue denotes the second highest in each column 
\end{table*}
\renewcommand\arraystretch{1}

\renewcommand\arraystretch{1.2}
\begin{table*}[t]
\small
\centering
\caption{Methods Tested Using the N3D-ReID framework \cite{aishahnot3d}} \label{tab:BigtoSmallN3D}
\begin{tabular}{l|ccc|ccc|}
\hline\hline
Backbone & \multicolumn{3}{c|}{Temporally-Near} & \multicolumn{3}{c|}{Big-to-Small} \\ \hline
 & mAP & rank-1 & rank-5 & mAP & rank-1 & rank-5 \\ \hline
ResNet-18 & 74.3 & 67.7 & 71.0 & 36.4 & 25.8 & 29.0 \\
ResNet-34 & 70.1 & 64.5 & 67.7 & 37.8 & 29.0 & 32.3 \\
ResNet-50 & 79.5 & {\textit{\color{blue}74.2}} & 77.4 & 38.5 & 29.0 & 32.3\\
SE-ResNet-50 & 72.1 & 64.5 & 71.0 & 40.2 & {\textit{\color{blue}32.3}} & 35.5\\
SE-ResNeXt-50 & 72.0 & 67.7 & 67.7 & 39.4 & 29.0 & 35.5\\
ViT Small & 79.2 & 71.0 & 77.4 & 39.6 & 29.0 & 32.3\\
ViT Base & 77.0 & 71.0 & 77.4 & 41.6 & 29.0 & 38.7\\
ResNet50mid & 78.7 & 71.0 & 77.4 & {\color{red}45.6} & {\color{red}35.5} & {\color{red}41.9}\\
OSNet & {\color{red}81.5} & {\color{red}77.4} & {\color{red}80.7} & 35.2 & 22.6 & 29.0\\
MLFN & 74.3 & 67.7 & 71.0 & 40.8 & {\textit{\color{blue}32.3}} & {\color{red}41.9}\\
PCB & {\textit{\color{blue}80.5}} & {\textit{\color{blue}74.2}} & {\color{red}80.7} & 39.3 & 29.0 & 32.3\\
HACNN & 74.1 & 67.7 & 74.2 & 41.6 & {\textit{\color{blue}32.3}} & 35.5\\
IBN-A & 72.0 & 64.5 & 67.7 & {\textit{\color{blue}41.9}} & {\textit{\color{blue}32.3}} & 35.5\\
\hline\hline                      
\end{tabular}
\\\medskip
Red denotes the (joint) highest in each column, blue denotes the (joint) second highest in each column 
\end{table*}
\renewcommand\arraystretch{1}
\subsection{Evaluation Protocol}\label{sec:Protocol}

We use the standard mean average precision (mAP), and rank based metrics to evaluate the selected state-of-the-art methods. The test set is split into a \emph{query set} and a \emph{gallery set}, with 31 identities each. Given a query image, $q$, the Re-ID framework ranks all gallery images, $g_i$ in order of likelihood that $g_i = q$, i.e. they contain the same UAV.

The rank-$r$ matching rate is the percentage of query images with a positive gallery image within the highest $r$ ranks.
The precision at rank $r$, $P_r$, compares the number of true positives (TP) with the total number of positives in the top $r$ ranks:
\begin{equation}
    P_r = \tfrac{\mathrm{TP}}{\mathrm{TP + FP}}\com
\end{equation}
where FP is the number of false positives.
As we only have one gallery image per query image, the mAP is calculated via
\begin{equation}
    \operatorname{mAP} = \frac{1}{Q}\sum_{q=1}^Q \frac{1}{r_q} \com
\end{equation}
where the correct identity of $q$ is found at rank $r_q$, and $Q$ is the total number of query images.

All experiments were performed using the torchreid framework \cite{zhou2019torchreid} on an NVIDIA RTX 2080 Ti GPU. All backbones were pre-trained on ImageNet.

\section{Evaluation}
We conduct an extensive benchmark evaluation over both the Temporally-Near and Big-to-Small re-identification challenges.

\subsection{Results}
Results on the `Temporally-Near' and  `Big-to-Small' sub-challenge dataset splits can be found in Table \ref{tab:TempNear} and \ref{tab:BigtoSmall}, respectively. ViT Base with CE+Triplet loss comprehensively outperforms all other methods on the Big-to-Small sub-challenge, and has fourth highest mAP on the Temporally-Near sub-challenge.
From Figure \ref{fig:Ranks}, rows two and three, we observe that ViT returns a similar ranking list on query UAVs that have different colour. It follows that ViT is capturing shape information as well as colour, which we hypothesise is due to its global self-attention mechanism, yielding superior performance compared to convolutional methods that rely on a local receptive field. This is in-keeping with the results of \cite{issacmedina21unmanned}, which corroborates the suitability of transformer networks towards detecting and identifying small objects such as drones. Similar to ViT, PCB also splits the input image into parts and obtains good performance across both tasks. This indicates that a part-based strategy can be effective for UAV Re-ID.

As expected, Big-to-Small is more challenging than Temporally-Near due to the extreme variation in scale. The best rank-1 matching rate of 77.4\% from generic architectures such as ResNet-18 and ViT is a strong baseline under the Temporally-Near sub-challenge. For real-world tracking systems, Re-ID is performed with only a few possible matches, rather than the entire test data set. These methods should therefore be sufficiently strong to be immediately employed within real-world systems. 

In contrast, Big-to-Small has top rank-1 and rank-5 matching rates of just 35.5\% and 54.8\%, respectively. We can attribute the difficulty of the challenge to the reduced colour and structure detail available to networks at a small scale, limiting the number of a differentiating features to identify. While colour exists, `blocky' compression artifacts are much more prevalent and there is very little variation across the image. As such, networks must be capable of identifying UAV from low-quality shape information, which only a few networks are capable of doing at this scale. Although ViT demonstrates potential in this regard, this sub-challenge requires further research to develop UAV-specific architectures sufficiently robust to scale and pose, and thus able to identify far away UAV.


The networks specific to Re-ID generally do not perform as well as generic networks. One reason for this is that extensive hyperparameter tuning is performed on generic networks to maximise classification performance on ImageNet, with a huge variety of objects seen. ReID-specific networks, although pre-trained on ImageNet, tune hyperparameters to maximise performance on person Re-ID data sets. Having specialised on humans, they have less functional ability to be transferred to different objects. However, PCB, which uses a ResNet-50 backbone (optimised for ImageNet), does still attain strong performance. 

In almost all cases, cross-entropy loss performance exceeds triplet loss. Further, the combined loss is occasionally unable to yield higher performance than cross-entropy alone. It is a common occurrence however, that triplet loss performance improves as the number of classes within the data set increases. Furthermore, because UAV-reID only allows one-to-one matching, we cannot harness the power of hard-positive mining. We expect that triplet loss will generate better results, and perhaps exceed cross-entropy, when a more comprehensive data set is made available.

The results from the Not-3D Re-ID framework (Table \ref{tab:BigtoSmallN3D}) corroborate our findings. Indeed, the additional loss functions incorporated into one multi-loss aggregation function are generally unable to improve results, but instead offer comparable results (+/-1\%) over the earlier loss formulations (Table \ref{tab:TempNear}, \ref{tab:BigtoSmall}). This is again perhaps attributable to the lack of effective hard-positive mining and few available classes. We can once again conclude that complex state-of-the-art person re-identification networks are less suited to UAV re-identification than shallower, simpler networks. In this regard, we can firstly observe that the IBN-A network does not out-perform the other networks in either challenge. The mAP performance of IBN-A under the temporally-near challenge (72.0\%) is significantly inferior to other backbone architectures. Secondly, the N3D-ReID framework is only able to improve upon ResNet50 (79.5\% mAP over 76.7\% mAP) and ViT Small (79.\% mAP over 75.6\% mAP) generic re-identification networks. However, N3D-ReID yields consistently stronger results for the Re-ID specific networks under the Temporally-Near challenge, with the exception of HACNN (74.1\% mAP compared to 77.7\% mAP).

Overall, OSNET performs the strongest with the N3D-ReID configuration, achieving 81.5\%. However, this does not improve upon ResNet-18 with just cross entropy loss (81.9\%, Table \ref{tab:TempNear}). Any improvements upon the Big-to-Small challenge results are similarly negligible when employing N3D-ReID. ResNet50mid generates the highest mAP of 45.6\% in this regard, less than that of ViT Base, 46.5\%, when using a combination of only cross entropy and triplet loss. Nevertheless, the results are further indicative that networks that achieve good results on the Temporally-Near challenge are not necessarily well-suited for the Big-to-Small challenge; the best performing networks under the N3D-ReID framework for Temporally-Near (OSNet, PCB, ResNet50) are disjoint from those suited to Big-to-Small (ResNet50mid, ViT Base, MLFN).


\subsection{Interpreting Vision Transformers}
\begin{figure}[t]
\centering
{\includegraphics[width=\linewidth]{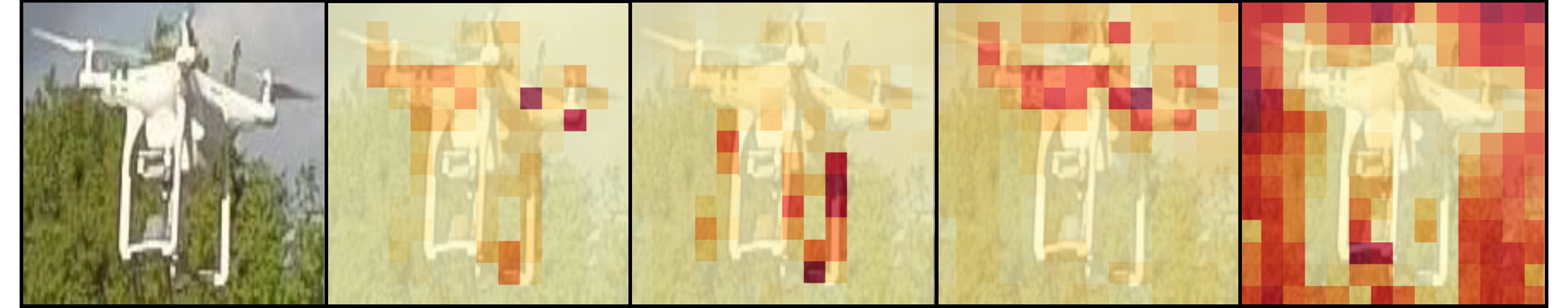}}
\caption{Attention visualisation of the transformer mechanism within ViT on the Big-to-Small setting. Attention from four different heads of the CLS token is presented. Different attention heads attend to different parts of the image, forming a more robust feature representation.} \label{fig:uav_global_attn}
\end{figure}

Across all experiments, ViT attains the highest performance on the Big-to-Small challenge with 46.5\% mAP. We visualise the attention maps to get a better understanding of how they achieve this. Figure \ref{fig:uav_global_attn} is a visualisation of four different attention heads of the CLS token.

The first attention map attends to the entire UAV, the second attends to its legs, the third to the propellers and the top of the UAV. This demonstrates clearly how it is encoding features and what the final feature representation consists of. The fourth attention map isolates the background. Even though the background is complicated, the attention head identifies that the drone is the foreground object, and considers the clouds and the trees together. This gives confidence that ViT has a good understanding of the image, and that the feature representation is composed in a structurally sound manner.



\begin{figure}[t]
\centering
{\includegraphics[width=0.8\linewidth]{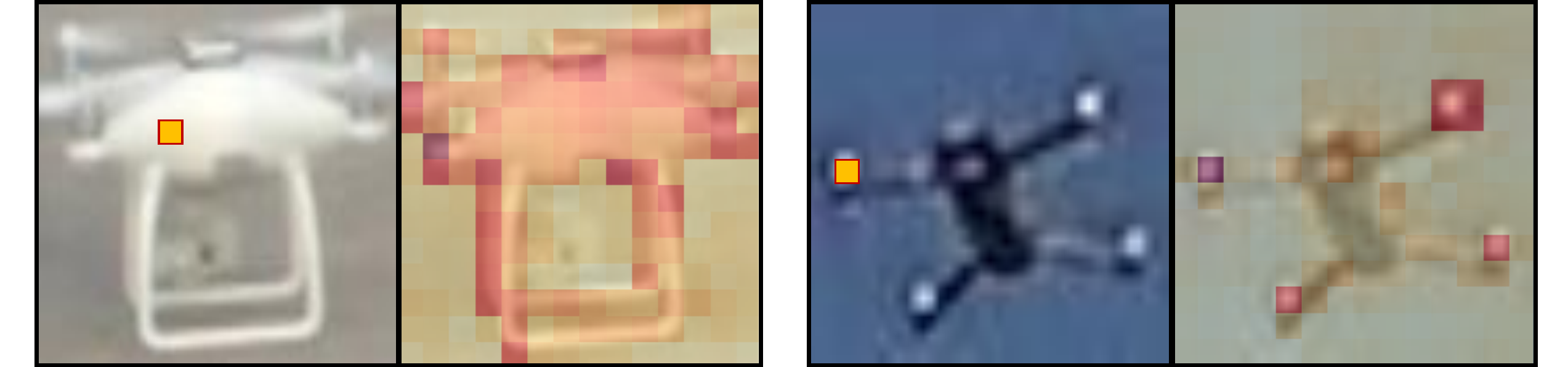}}
\caption{Attention visualisation of the transformer mechanism within ViT on the Big-to-Small setting. The query patch is indicated by a yellow square.} \label{fig:uav_local_attn2}
\end{figure}

Figure \ref{fig:uav_local_attn2} visualises attention from a specific image patch, indicated via the yellow box. On the left, the query patch occurs on the UAV, and the resulting attention strongly segments the UAV from the background. On the right, the query patch occurs on one of the propellers, and the attention head attends to each of the other propellers. One of the advantages of transformers over traditional convolution is their ability to learn non-local relationships between image patches to obtain a stronger feature representation. These visualisations demonstrate this process in action.

\section{\uppercase{Conclusions}}\label{sec:conc}
We have proposed the challenge of UAV re-identification and performed a benchmark study to examine the effectiveness of a variety of deep learning techniques. Vision transformers trained with a combined cross-entropy and triplet loss attain strong performance across both tasks, achieving the highest mAP on the Big-to-Small challenge and the 4th highest mAP on the Temporally-Near setting. A range of methods can re-identify UAVs over a short time period with high precision. Of these methods, ResNet-18 (mAP 81.9 \%) appears to be easiest to fit into tracking frameworks due to its high performance and relatively small model size.

Although the Big-to-Small data set split is very challenging, vision transformers have shown great promise with respect to handling extreme scale transformation. We can attribute this behaviour to their superior performance over other architectures due to their ability to learn relationships between distant image patches.

There is clear motivation for future work. A large multi-view UAV Re-ID data set with more instance classes would be beneficial to get the full potential out of deep networks and multiple loss functions. Based on its success in this benchmark, we also wish to develop an improved vision transformer by incorporating techniques used in convolutional neural networks to handle scale changes, such as concatenating outputs from different layers. Nevertheless, our work establishes a clear baseline for UAV re-identification performance, of which the benefits are evident within potential UAV tracking frameworks.




\section*{\uppercase{Acknowledgements}}
This work is funded in part by the Future Aviation Security Solutions (FASS) programme, a joint initiative between DfT and the Home Office (Ref: 007CD). 
This work is made possible by the WOSDETC dataset.

\bibliographystyle{apalike}
{\small\bibliography{reid}}

\end{document}